\documentclass[review]{elsarticle}

\usepackage{graphicx} % Required for inserting images
\usepackage{multirow}
\usepackage{tabularx}
\usepackage{booktabs}

\journal{Journal of \LaTeX\ Templates}

\newcounter{example}

\usepackage{afterpage}
\usepackage{capt-of}
\begin{document}

\begin{frontmatter}

\title{Deep Learning Detection Method for Large Language Models-Generated Scientific Content }

\author[psut]{Bushra Alhijawi \corref{cor1}}
\ead{b.alhijawi@psut.edu.jo}
\cortext[cor1]{Corresponding author}

\author[psut]{Rawan Jarrar}
\ead{raw20228063@std.psut.edu.jo}

\author[psut]{Aseel AbuAlRub}
\ead{ase20228038@std.psut.edu.jo}

\author[psut]{Arwa Bader}
\ead{Arw20228003@std.psut.edu.jo}

\address[psut]{Princess Sumaya University for Technology, Amman, Jordan}

\begin{abstract}
Large Language Models (LLMs), such as GPT-3 and BERT, reshape how textual content is written and communicated. These models have the potential to generate scientific content that is indistinguishable from that written by humans. Hence, LLMs carry severe consequences for the scientific community, which relies on the integrity and reliability of publications. This research paper presents a novel ChatGPT-generated scientific text detection method, AI-Catcher. AI-Catcher integrates two deep learning models, multilayer perceptron (MLP) and convolutional neural networks (CNN). The MLP learns the feature representations of the linguistic and statistical features. The CNN extracts high-level representations of the sequential patterns from the textual content. AI-Catcher is a multimodal model that fuses hidden patterns derived from MLP and CNN. In addition, a new ChatGPT-Generated scientific text dataset is collected to enhance AI-generated text detection tools, AIGTxt. AIGTxt contains 3000 records collected from published academic articles across ten domains and divided into three classes: Human-written, ChatGPT-generated, and Mixed text. Several experiments are conducted to evaluate the performance of AI-Catcher. The comparative results demonstrate the capability of AI-Catcher to distinguish between human-written and ChatGPT-generated scientific text more accurately than alternative methods. On average, AI-Catcher improved accuracy by 37.4\%.

\end{abstract}


\end{frontmatter}

%\linenumbers

\section{Introduction}

Information and communication technology (ICT) advancements enable easy access to a variety of content (e.g., text, images, and videos). Hence, ICT facilitates reaching several research and information sources. However, plagiarism has also become a severe problem for publishers, researchers, and educational institutions due to ICT \cite{maurer2006plagiarism}. Plagiarism is the act of using someone else's work as one's own, without reference to the original source \cite{lukashenko2007computer}. Commonly, plagiarism includes copying textual information and ideas, paraphrasing, misinformation of references, and using codes with permission \cite{maurer2006plagiarism}. Today, scientific and academic plagiarism is widespread in the scientific community \cite{zouaoui2022multi}. Academic plagiarism has several consequences:
\begin{itemize}
\item False findings can spread and influence subsequent research or practical applications \cite{gipp2014citation}.
\item Wastes resources where identifying, reviewing, and penalizing plagiarized research requires high effort from the reviewers, affected institutions, and funding organizations \cite{wager2014defining}.
\item Funding agencies may award grants for plagiarized ideas or accept plagiarized research works as the outcomes of research projects \cite{foltynek2019academic}.
\end{itemize}

\begin{figure} [h!]
\begin{center}
\includegraphics[scale=0.50]{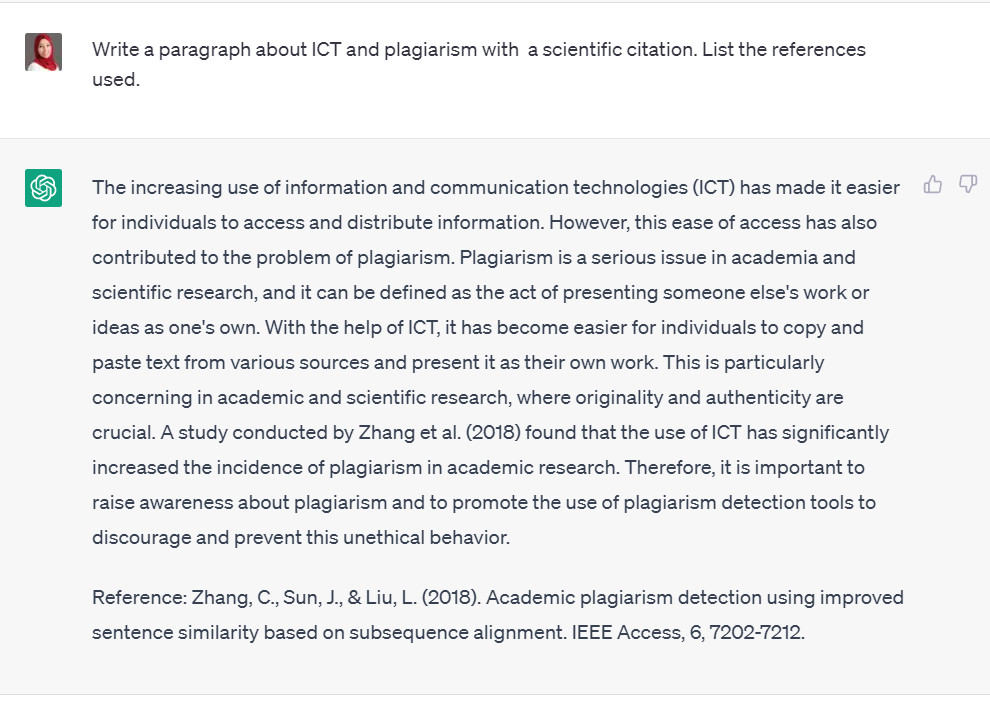}
\end{center}
\caption{Example of ChatGPT-generated scientific text.}
\label{gptEx1}
\end{figure}

Artificial intelligence (AI) technologies have recently been employed as powerful tools to improve productivity in various fields. These technologies can generate undetectable research works by humans and automated technologies \cite{elali2023ai}. Therefore, AI-based text generation technologies transform the plagiarism issue into a more critical problem. The fabrication of research works has severe consequences for the scientific community that depends on the integrity and reliability of these publications to make decisions about changes in various fields, including sociology, economics, politics, and medicine \cite{elali2023ai}. In November 2022, OpenAI launched a new robust language model chatbot called Chat Generative Pre-trained Transformer (ChatGPT) \footnote{Released by OpenAI in November 2022. https://chat.openai.com/chat}. Although other AI chatbots are in use, ChatGPT has been revolutionary for many. ChatGPT readily fabricates research works that easily bypass plagiarism detection tools \cite{elali2023ai}. ChatGPT has been listed as a co-author for several published research articles \cite{o2022open,transformer2022rapamycin}. Figure \ref{gptEx1} shows an example of ChatGPT-generated scientific text. The generated text can be used as a paragraph in the introduction section of the plagiarism detection article. However, the ChatGPT-generated text in the example involves falsifying non-existent data and results, whereas the reference used does not exist. 

These issues motivated us to propose a novel method to detect ChatGPT-generated scientific text. The contribution of this study involves the development and implementation of a new ChatGPT-generated scientific text detection method, namely, AI-Catcher. In addition, a new dataset is collected to train and develop the AI-Catcher, called the AI-Generated Text (AIGTxt) dataset. AI-Catcher is a shared-learning model that fuses hidden patterns extracted from multilayer perceptron (MLP) and convolutional neural networks (CNN). The proposed method prepares a set of linguistic and statistical features from the scientific text during the preparation phase of the MLP. CNN learns the sequential patterns from the textual content. To the best of our knowledge, AI-Catcher is the first model that can distinguish between human-written and ChatGPT-generated scientific text. In addition, the AIGTxt dataset is the first public dataset that includes ChatGPT-generated scientific texts.

The main objective of this research is to design and develop an AI-generated scientific textual content detection method. The contributions of this research paper are:

\begin{itemize}
\item A new ChatGPT-generated scientific text dataset, AIGTxt. AIGTxt is the first dataset designed to enhance AI-generated scientific text detection tools. 

\item A novel ChatGPT-generated scientific text detection method using deep learning and natural language processing (NLP), AI-Catcher. AI-Catcher is the first model that aims to identify academic scientific text generated using ChatGPT.
\end{itemize}

The remainder of this paper is organized as follows: Section \ref{LR} presents a summary of the recently published works on plagiarism detection, machine-generated text detection, and authorship identification methods. Section \ref{data} describes the AIGTxt dataset. Section \ref{gptCS} details the proposed method, AI-Catcher. Section \ref{exp} focuses on evaluating the proposed method and reporting the comparison results with alternative methods. Section \ref{con} concludes the research and presents potential future directions.

\section{Literature Review}
\label{LR}

% READ IT 
% https://www.science.org/doi/full/10.1126/science.adg7879
% https://www.nature.com/articles/d41586-023-00288-7
% https://arxiv.org/pdf/2011.01314.pdf
% https://ieeexplore.ieee.org/stamp/stamp.jsp?tp=&arnumber=10177704

Recent advances in large language models (LLMs) have demonstrated a quantum leap in various NLP tasks. LLMs significantly improve the performance in multiple tasks, including generating human-like texts. However, several ethical concerns have to be considered and addressed, such as plagiarism, false information, educational concerns, and research work fabrication. Therefore, substantial research efforts have been made to develop plagiarism detection \cite{alsallal2019integrated,meuschke2018hyplag,meuschke2018adaptive,gharavi2016deep,vysotska2018defining,altheneyan2020automatic,arabi2022improving,veisi2022multi,alvi2021paraphrase,malandrino2022adaptive,Yalcin2022an}, machine-generated text detection \cite{saravani2021automated,gambini2022pushing,sarin2020convgrutext,adelani2020generating,bao2019learning,guo2023accurate,An2023use,Sadiq2023Deepfake,Chen2023STADEE,fagni2021tweepfake}, and authorship identification methods \cite{alsallal2019integrated,vysotska2018defining,uchendu2020authorship,uchendu2023attribution}. In addition, several commercial tools for detecting ChatGPT-generated text are currently available online \cite{ZeroGPT,GPTZero,Writer}.

This section reviews the recent contributions to the research problem, plagiarism detection methods (Section \ref{pd}), machine-generated text detection methods (Section \ref{machG}), and authorship identification methods (Section \ref{AuthI}). Also, Section \ref{com} reviews popular commercial tools for detecting ChatGPT-generated text. In addition, Section \ref{dist} summarizes the objectives and contributions of this research, along with distinctions from other approaches.

\subsection{Plagiarism Detection Methods}
\label{pd}

Identifying content that has been copied/paraphrased from other sources without proper citation is known as plagiarism detection. Gharavi et al. \cite{gharavi2016deep} designed a plagiarism detection method using Word2Vec. Their approach aggregates multi-dimensional word vectors to represent sentences. Each sentence in an input document is compared with the source document using cosine and Jaccard similarities. Alsallal et al. \cite{alsallal2019integrated} developed a plagiarism detection method based on the statistical characteristics of the text and latent semantic analysis (LSA). Their technique extracts a set of features that model the author's writing style. MLP was used to build an authorship identification model. Meuschke et al. \cite{meuschke2018hyplag} presented a hybrid plagiarism detection approach for academic documents called HyPlag. HyPlag identifies potentially suspicious research content by analyzing 
mathematical equations, images, textual content, and citations. Alvi et al. \cite{alvi2021paraphrase} designed a paraphrase type identification technique using contexts and word embeddings for plagiarism detection. Meuschke et al. \cite{meuschke2018adaptive} proposed an adaptive image-based plagiarism detection technique. They developed new image similarity measures called ratio hashing and position-aware OCR text matching and integrated them with image analysis methods. Vysotska et al. \cite{vysotska2018defining} designed a plagiarism detection method by analyzing the author's style for scientific articles. Their method extracts linguistic features to define the author's style and uses a statistical classifier for authorship identification. Altheneyan et al. \cite{altheneyan2020automatic} proposed a support vector machine (SVM)-based plagiarism detection approach. Their method exploits the obfuscated text's lexical, syntactic, and semantic features to train SVM to identify whether two given sentences are plagiarized. Arabi and Akbari \cite{arabi2022improving} developed an extrinsic plagiarism detection approach based on structural and semantic hybrid similarity of sentences. Their method uses a word weighting technique (e.g., TF-IDF and WordNet ontology) integrated with FastText to form a sentence structural matrix. Veisi et al. \cite{veisi2022multi} presented a multi-level text document similarity for plagiarism detection. They estimated the similarity between the input and source documents using the cosine distance between word vectors. Malandrino et al. \cite{malandrino2022adaptive} developed an adaptive meta-heuristic method for music plagiarism detection. The meta-heuristic integrates a text similarity-based method and a clustering-based method. Yalcin et al. \cite{Yalcin2022an} proposed a plagiarism detection technique depending on part-of-speech (POS) tag n-grams and word embedding. The PSO tag n-grams help measure the syntactic similarities between source and suspicious sentences. The word embedding (i.e., Word2Vec) and longest common subsequence measure are used to compute the semantic relatedness between words in the source and suspicious sentences. 

% Arabi and Akbrai \cite{Arabi2022Improving} presented a hybrid weighted similarity measure to improve plagiarism detection in text documents. They developed two methods to identify extrinsic plagiarism. The first method detects the similarity between suspicious documents using FastText and TF-IDF, while the second method employs WordNet ontology and TF-IDF to compute the similarity between suspicious documents.

\subsection{Machine-generated Text Detection Methods}
\label{machG}

Several approaches have been used to develop machine-generated text detection methods for social media posts \cite{saravani2021automated,gambini2022pushing,sarin2020convgrutext,Sadiq2023Deepfake,fagni2021tweepfake}, product reviews \cite{adelani2020generating,sarin2020convgrutext}, news \cite{bao2019learning,guo2023accurate,sarin2020convgrutext,Chen2023STADEE}, and student essays \cite{An2023use}. The machine-generated text is produced using GPT-2 \cite{saravani2021automated,adelani2020generating,bao2019learning,guo2023accurate,Sadiq2023Deepfake,fagni2021tweepfake} and GPT-3 \cite{gambini2022pushing,An2023use,Chen2023STADEE}. Saravani et al. \cite{saravani2021automated} proposed a new method for bot detection that focuses on the textual content of posts. They used a CNN model to distinguish between human-generated and deep fake text. Adelani et al. \cite{adelani2020generating} employed BERT to detect the fake online reviews that are generated using GPT-2. Their method analyzes the sentiment of the reviews to detect reviews with undesired sentiments. Gambini et al. \cite{gambini2022pushing} developed an ensemble method to detect GPT-3 social media texts based on lexical, syntactic, and topical features. Bao et al. \cite{bao2019learning} proposed an SVM-based fake text detection method that analyzes the semantic coherence of the text. Guo et al. \cite{guo2023accurate} developed a BERT-based fake news detection method. Their method employs a deep layer-wise relevance propagation to track the importance of individual features in a deep neural network. Sarin and Kumar \cite{sarin2020convgrutext} employed convolutional gated recurrent units to detect fake tweets. An et al. \cite{An2023use} developed a ChatGPT-generated essay detection method using a similarity score. They employed student-written essays in response to eight prompts. These prompts are used to generate ChatGPT-generated essays. Sadiq et al. \cite{Sadiq2023Deepfake} integrated CNN and FastText word embedding to develop a machine-generated tweet identification method. Chen and Liu \cite{Chen2023STADEE} integrated statistical features of text with a sequence-based deep classification method to identify ChatGPT-generated news. Several statistical features are extracted, such as token cumulative probability, token rank, and the information entropy of the distribution at each position.

% https://aclanthology.org/2022.naacl-main.88/
% % https://arxiv.org/abs/2305.12519
% https://journals.plos.org/plosone/article?id=10.1371/journal.pone.0251415
% https://peerj.com/articles/cs-443/
% https://arxiv.org/abs/2109.04825
% https://arxiv.org/abs/2305.07969

% MORE tokens 
% https://ieeexplore.ieee.org/abstract/document/9983964?casa_token=2-gT3wMvlwsAAAAA:ny8IbwKAtnCyl5EwXgGoKe1aImZcNeNb-KSZmn4a-nEgJsMP2HBjYdBINh_WUYUk42wV_opHXlZHTw

\subsection{Authorship Identification Methods}
\label{AuthI}

Authorship identification (i.e., authorship attribution) employs computational linguistics and NLP to determine the author of a given text based on linguistic features, writing style, and other characteristics unique to that individual. Numerous researchers have addressed plagiarism detection as an authorship identification challenge \cite{vysotska2018defining,uchendu2020authorship,uchendu2023attribution,alsallal2019integrated}. Vysotska et al. \cite{vysotska2018defining} developed a plagiarism detection method in scientific articles by analyzing the author's style. They trained the Bayesian classifier based on quantitative characteristics of the author's style. Uchendu et al. \cite{uchendu2020authorship} investigated the differences between the linguistic features of human-written and machine-generated text. They developed deep learning methods, i.e., RNN, CNN, and embedding, to detect whether two texts are generated by the same method, if the text is written by a human or machine, and identify the author of a text. Alsallal et al. \cite{alsallal2019integrated} proposed a plagiarism detection method without reference collection. Their method adopts latent semantic analysis to extract the most common word usage patterns.

\subsection{Commercial ChatGPT Detection Tools }
\label{com}

Recently, several commercial ChatGPT detection tools are available online for users, such as ZeroGPT \cite{ZeroGPT}, GPTZero \cite{GPTZero}, and Writer \cite{Writer}. These tools provide businesses, organizations, and individuals APIs to protect themselves from the risks associated with AI-generated content, such as plagiarism, misinformation, and spam. ZeroGPT \cite{ZeroGPT} uses ML algorithms and NLP techniques to analyze text and identify patterns common in AI-generated text. GPTZero \cite{GPTZero} analyzes the textual data to determine the AI-generated content. Their approaches depend on two features: perplexity and burstiness. Perplexity measures how difficult it is to predict the next word in a sequence of words. The higher the perplexity score, the more complex and unpredictable the text is, therefore, more likely to be human-written. Burstiness measures how evenly distributed the perplexity scores are throughout a text. A text with a lower burstiness score is more likely to be AI-generated. Writer \cite{Writer} works by analyzing a text for patterns and characteristics common in AI-generated text.

% https://dl.acm.org/doi/abs/10.1145/3606274.3606276

\subsection{Objectives and Distinctions}
\label{dist}

This section presents the objectives and the differences between the proposed AI-generated content detection method, AI-Catcher, and other related approaches. 

We categorized the related works according to the addressed problem into three types: (1) plagiarism detection, (2) machine-generated text detection, and (3) authorship identification methods. Plagiarism detection methods focus on identifying copied content without proper citation. These methods are developed using ML, semantic analysis, and meta-heuristic techniques. The reviewed machine-generated text detection approaches aim to detect textual content for social media posts, product reviews, news, and student essays using GPT-2 and GPT3. Different deep learning algorithms are used to develop AI-generated text detection methods, such as CNN and transforms (e.g., BERT). The authorship identification methods focus on determining the author of a given text. The reviewed articles employed deep learning (e.g., CNN and RNN) and semantic analysis methods. In addition, three popular commercial tools for detecting the textual content generated using ChatGPT are reviewed (i.e., ZeroGPT, GPTZero, and Writer). These tools provide individuals and businesses with free and paid services to identify ChatGPT-generated content.  

The main objective of this research is to design and develop an AI-generated scientific textual content detection method. To the best of our knowledge, we have not found any work in the AI-generated text detection method field in the literature that focuses on identifying the academic scientific textual content (i.e., research papers) generated using LLMs, such as ChatGPT. The proposed method, AI-Catcher, employs deep learning and NLP to detect ChatGPT-generated scientific textual content.

\section{AI-Generated Text Dataset}
\label{data}

\begin{figure} [h!]
\begin{center}
\includegraphics[scale=0.5]{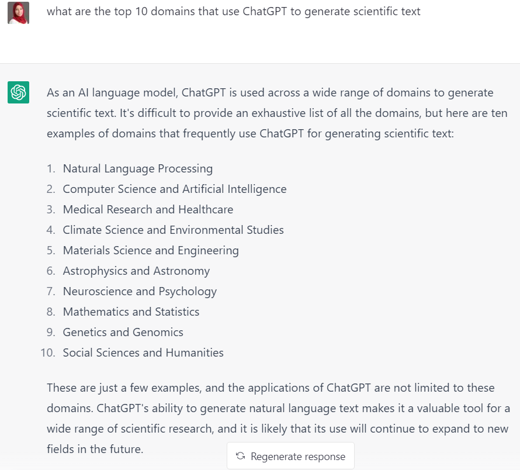}
\end{center}
\caption{Top-10 domains that use ChatGPT to generate scientific text.}
\label{top_10}
\end{figure}

AI-Generated Text (AIGTxt) dataset is a new ChatGPT-generated scientific text dataset designed to enhance AI-generated text detection tools. AIGTxt contains scientific texts collected from published academic articles across ten different domains. The ten different domains are selected based on the answer of ChatGPT for "What are the Top-10 domains that use ChatGPT to generate scientific text" (Figure \ref{top_10}). All selected research papers were published before 2022 to ensure that ChatGPT was not used to write the scientific content. 

\begin{figure} [h!]
\begin{center}
\includegraphics[scale=0.45]{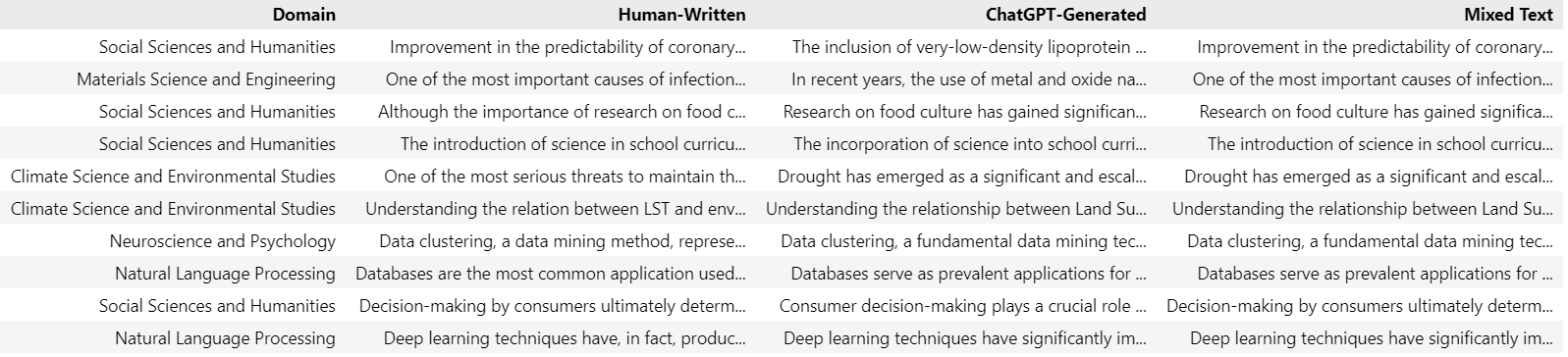}
\end{center}
\caption{Sample of AIGTxt dataset.}
\label{dataSample}
\end{figure}

AIGTxt includes three classes: human-written, ChatGPT-generated, and mixed text. In total, AIGTxt includes 1,000 human-written texts with their corresponding ChatGPT-generated and mixed text. Figure \ref{dataSample} shows a sample of AIGTxt. The process of generating AIGTxt is detailed below.

\begin{enumerate}
\item Human-written text collection. One to three paragraphs from the introduction or literature review sections are manually collected. The collected text is labeled as human-written text.

\item  ChatGPT-generated text generation. Manual interaction with ChatGPT is orchestrated to rewrite the original human-written text. Varied queries are used during this stage, such as "Rewrite the following paragraphs more professionally and with an IEEE scientific citation:" and "Revise, restructure, and rewrite the following paragraph, maintaining a professional tone". The collected text is labeled as ChatGPT-generated text. Figure \ref{ex1} shows an example of ChatGPT-generated text. 

\begin{figure} [h!]
\begin{center}
\includegraphics[scale=0.6]{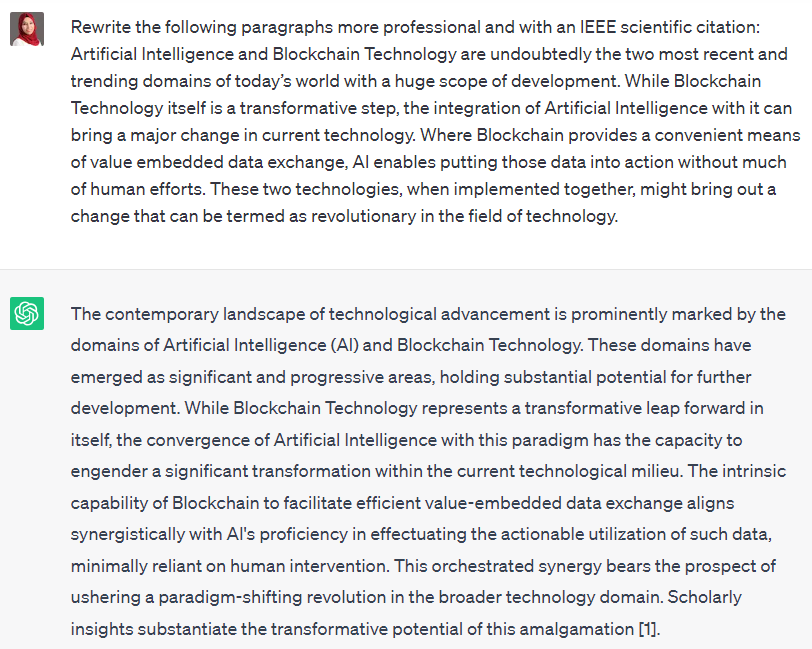}
\end{center}
\caption{Example of ChatGPT-generated text.}
\label{ex1}
\end{figure}

\item Mixed text generation. A harmonious fusion of human-written and ChatGPT-generated text is manually and meticulously blending 50\% of each class. The collected text is labeled as mixed text.

\end{enumerate}

Table \ref{allS} summarizes the characteristics of AIGTxt.  AIGTxt comprises 1000 records per class (i.e., Human, ChatGPT, and Mixed), distributed across ten distinct topics. The average length of the scientific text is approximately 194 words, while the dataset encompasses a rich vocabulary of 29,226 unique words. It is worth mentioning that the number of unique words is computed without applying any preprocessing steps except excluding the stopwords and citations. Table \ref{domS} presents a statistical summary of AIGTxt per topic. The domain with the maximum number of records (510) in AIGTxt is the Social Sciences \& Humanities domain. In contrast, the domain with the minimum number of records (45) in AIGTxt is the Mathematics \& Statistics domain. Genetics and genomics scientific text has the maximum average text length (246.80 words) among all considered fields. On the other hand, the Astrophysics \& Astronomy domain has the minimum average text length (160.72 words) among all considered domains. The AIGTxt dataset will be available for download on the GitHub platform and updated regularly.

\begin{table}[h!]
\centering
\caption{Statistical Summary of AIGTxt Dataset}
\begin{tabular}{|l|c|}
\hline
\textbf{Measure} & \textbf{Value} \\ \hline
Total number of observations & 3000\\ \hline
Number of records per class & 1000 \\ \hline
Number of classes & 3 \\ \hline
Number of topics & 10 \\ \hline
Average paragraph length (word) & 193.49 \\ \hline
Number of unique words (excluding stopwords) & 29226 \\ 
\hline
\end{tabular}
\label{allS}
\end{table}

\begin{table}[h!]
\centering
\caption{Statistical Summary of AIGTxt Dataset per Domain}
\label{domS}
\small 
\begin{tabularx}{\textwidth}{@{}lXXXXXX@{}}
\toprule
Domain & Num of Records  & Avg Paragraph Len & Min Paragraph Len & Max Paragraph Len & Unique Words \\
\midrule
Astrophysics \& Astronomy & 432 & 160.72 & 66 & 425 & 6806 \\
Climate Science \& Environmental Studies & 315 & 215.67 & 100 & 490 & 7114 \\
Computer Science \& Artificial Intelligence & 432 & 206.97 & 97 & 435 & 8310 \\
Genetics \& Genomics & 192 & 246.80 & 97 & 501 & 5190 \\
Materials Science \& Engineering & 144 & 179.19 & 97 & 312 & 3021 \\
Mathematics \& Statistics & 45 & 201.66 & 118 & 333 & 1516 \\
Medical Research \& Healthcare & 264 & 214.96 & 112 & 445 & 5576 \\
Natural Language Processing & 453 & 190.13 & 67 & 443 & 7131 \\
Neuroscience \& Psychology & 213 & 206.80 & 63 & 346 & 5123 \\
Social Sciences \& Humanities & 510 & 165.71 & 70 & 453 & 8908 \\
\bottomrule
\end{tabularx} 
\end{table}

\section{AI-Catcher}
\label{gptCS}

\begin{figure}[h!]
\centering 
\includegraphics[scale=0.4]{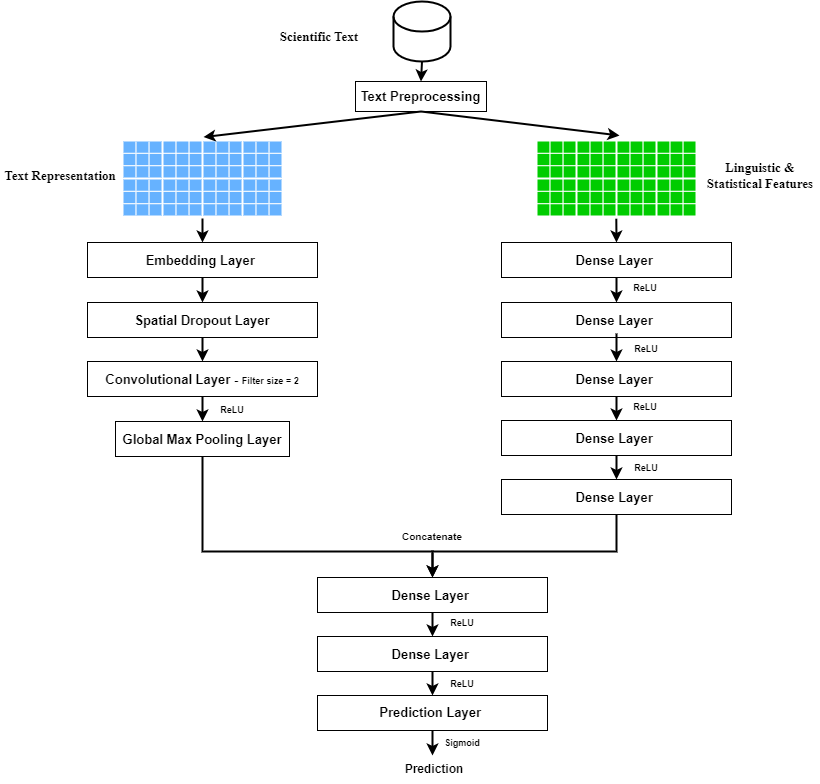}
\caption{Architecture of AI-Catcher}
\label{arc}
\end{figure}

AI-Catcher aims to identify whether a human or ChatGPT has written a particular scientific text. The proposed method addresses the machine-generated scientific text detection challenge as a binary classification problem. AI-Catcher is a multimodal model trained on two types of features: (1) linguistic and statistical features; and (2) textual content. Figure \ref{arc} shows the architecture of AI-Catcher. AI-Catcher framework comprises two phases:
\begin{itemize}
\item Data Preparation Phase (Section \ref{dp}).  AI-Catcher prepares the required linguistic features, statistical features, and text representation for the classification phase. 
\item Classification Phase (Section \ref{cp}).  AI-Catcher involves a multimodal model involving MLP and CNN components. The collaborative nature of the model empowers AI-Catcher to discriminate between human and Chat-GPT written scientific content effectively.
\end{itemize}

\subsection{Data Preparation Phase}
\label{dp}

AI-Catcher orchestrates distinct input feature preparation processes for each component of the classification model (i.e., MLP and CNN). The data preparation for the CNN component includes three processes: 
\begin{enumerate}

\item Text cleaning. AI-Catcher applies a lemmatization method to the text to reduce the words to their base or root form. Text cleaning provides AI-Catcher with several advantages: text normalization, reduction of data dimensionality, and the unification of different word forms into a singular entity.

\item Text encoding. AI-Catcher converts the cleaned text into a sequence of one-hot encoded vectors. This process assigns a unique integer index to each word in the text based on the specified vocabulary size.

\item Padding. AI-Catcher applies a padding process to the encoded text to provide a standardized representation. 

\end{enumerate}

AI-Catcher extracts thirteen linguistic and statistical features from the scientific text during the preparation phase of the MLP. The prepared features help AI-Catcher learn the text's complexity, richness, correctness, organization, coherence, and emotional tone. The extracted linguistic and statistical features are:
\begin{itemize}
\item Average sentence length: the average number of words in each sentence of a text. 
\item Average word length: the average number of characters in each text word.
\item Ratio of unique words: the ratio of distinct words to the total number of words in the text. 
\item Count of grammatical errors: the number of grammatical errors in the text. AI-Catcher uses a Python open-source grammar tool called LanguageTool \footnote{https://pypi.org/project/language-tool-python. Access Date: 26/8/2023} to identify the grammatical errors.
\item Count of discourse markers: the count of discourse markers in the text. A list of discourse markers in the English language (e.g., moreover, furthermore, and additionally) is determined to count discourse markers in the text. 
\item Ratio of nouns, pronouns, verbs, adverbs, and adjectives: the proportions of different parts of speech in the text.
\item Ratio of punctuation: the ratio of punctuation marks to the total number of words in the text. 
\item  Ratio of stopwords: the ratio of stopwords in the text. 
\item  Ratio of words with sentiment: the proportion of words in the text with an associated emotional sentiment. AI-Catcher employs the NLTK VADER sentiment analysis model \footnote{https://www.nltk.org/api/nltk.sentiment.vader.html. Access Date: 26/8/2023} to identify words with polarity scores greater or less than zero. 
\end{itemize}

\subsection{Classification Phase}
\label{cp}

The classification model of AI-Catcher consists of two components, MLP and CNN, each adept at extracting different forms of hidden patterns from different input features.

The input integer-encoded sequences of the CNN are transformed into dense vectors using the embedding layer. The embedding layer maps each word index to a corresponding vector representation to capture semantic relationships between words in the text. Later, the spatial dropout layer performs dropout on the dense word embeddings. This layer helps prevent overfitting by forcing the model to learn more robust representations by ensuring that no single word embedding dominates the learning process. The dropout layer randomly sets approximately 20\% of the embeddings to zero for each training iteration. The CNN component includes one convolutional layer connected with a global max pooling layer. The 1D convolutional layer captures local patterns in the form of bigrams within the textual data. The global max pooling performs max pooling across the sequence dimension of the convolutional output. Max pooling reduces the dimensionality of the convolutional layer outputs while retaining the most salient features. Therefore, the pooling layer emphasizes the most relevant information from each filter for the prediction process. 

The MLP component of AI-Catcher consists of five hidden layers, each composed of densely connected 256 neurons. Stacking multiple hidden layers enables the network to learn increasingly abstract and high-level features. Each hidden layer applies a linear transformation to its input, followed by the ReLU activation function. ReLU introduces the non-linearity to the MLP. Hence, AI-Catcher learns complex patterns and relationships within the data.

AI-Catcher is a multimodal model that fuses hidden patterns derived from MLP and CNN. This fusion occurs by combining two distinct types of feature representations:
\begin{itemize}
\item The feature representations of the linguistic and statistical features extracted by the MLP.
\item The high-level representations of the sequential patterns captured by the CNN.
\end{itemize}

The concatenation operation merges the different feature representations into a single feature vector. The concatenation layer aims to capture various aspects of the data, enhance feature representation, and learn from different feature types. Later, the combined feature vector is passed through two additional hidden layers to learn more abstract and high-level representation features for the human/ChatGPT classification task.

\section{Experiments and Results}
\label{exp}

This section details the evaluation of the AI-Catcher. Various experiments are conducted to compare AI-Catcher performance with that of alternative baseline classification methods and commercial AI-generated text detection tools. Section \ref{ed} details our experimental designs and setups. Sections \ref{rd} and \ref{dis} present and discuss the experimental results.

\subsection{Experiments Design}
\label{ed}

This section details the experimental setup of the proposed technique, AI-Catcher. Two different experiments are conducted to test the proposed ChatGPT-generated text detection model. The first experiment aimed to compare the results of AI-Catcher with those obtained from 

\begin{itemize}
\item Four baseline classification methods, i.e., logistic regression (LogReg), SVM, random forest (RandF), and Naive Bayes (NP).

\item Seven deep learning methods, i.e., CNN (AI-Catcher component), MLP (AI-Catcher component), MLP with two hidden layers (MLP2), MLP with a tower pattern of two hidden layers (MLP2-TP), MLP with a tower pattern of five hidden layers (MLP5-TP), charCNN \cite{zhang2015character}, and XLNet \cite{yang2019xlnet}. charCNN is a character-level convolutional network that is used for text classification, while XLNet is an autoregressive pre-trained language model. We used the implementations of Fagni et al. \cite{fagni2021tweepfake} for charCNN and XLNet.

\end{itemize}

The second experiment compared the AI-Catcher performance with three commercial detection tools, namely ZeroGPT \cite{ZeroGPT}, GPTZero \cite{GPTZero}, and Writer \cite{Writer}. 

Three experimental setups are considered: DEV1, DEV5, and DEV10. In DEV1, we partitioned the data into 80\% for training 
and 20\% for testing. In DEV5 and DEV10, the performance of the detection methods is evaluated using 5-fold cross-validation and
10-fold cross-validation, respectively. Four performance evaluation measures are used: accuracy, precision, recall, and F1-score. Accuracy (Eq. \ref{accu}) refers to the proportion of correctly predicted human-written and ChatGPT-generated text to the total count of predictions. Precision (Eq. \ref{prec}) quantifies the proportion of accurately predicted ChatGPT-generated text among all predicted observations. Recall (Eq. \ref{re}) quantifies the ratio of correctly predicted ChatGPT-generated text to the total number of actual ChatGPT-generated text. The F1-score (Eq. \ref{f1}) refers to the harmonic mean of the precision and recall metrics. All reported results in the research paper are obtained using the same dataset used for evaluating AI-Catcher (i.e., AIGTxt). The results of the commercial detection tools are obtained by
manual testing on the same 20\% testing dataset as utilized for AI-Catcher in the DEV1 setup. 

\begin{equation}
\label{accu}
Accuracy = \frac{TP + TN}{TP + FP + TN + FN}
\end{equation}

\begin{equation}
\label{prec}
Precision = \frac{TP}{TP + FP}
\end{equation}

\begin{equation}
\label{re}
Recall = \frac{TP}{TP + FN}
\end{equation}

\begin{equation}
\label{f1}
F1-score = 2 \times \frac{Precision \times Recall}{Precision + Recall}
\end{equation} 

where,

\begin{itemize}
    \item True Positive (TP) indicates the count of correctly predicted positive class instances.
    \item True Negative (TN) indicates the count of correctly predicted negative class instances.
    \item False Positive (FP) indicates the count of incorrectly predicted positive class instances.
    \item False Negative (FN) indicates the count of incorrectly predicted negative class instances. 
\end{itemize}

\subsection{Experimental Results}
\label{rd}

This section presents the results obtained by applying AI-Catcher on the AIGTxt dataset. The results are used to compare the proposed method with baseline classification methods, deep learning methods, and commercial detection tools. Table \ref{dev1res} and Table \ref{dev510res} show the results of the first experiment. For DEV1 (Table \ref{dev1res}), AI-Catcher achieves a significant improvement of 33.8\%, 31.92\%, 19.7\%, and 27.97\% in terms of accuracy, precision, recall, and F1-score, respectively. The accuracy improvements exhibited by the proposed method range from 2.65\% to 89.19\%. Significantly fewer errors are obtained using AI-Catcher than the baseline classification and deep learning methods in the DEV5 experimental setup. On average, the proposed method leads to (32.03\%,  34.72\%, 27.25\%, and 31.6\%) better accuracy, precision, recall, and F1-score than alternative methods, respectively. On average, the comparative results showed (34.86\%, 134.97\%, 65.52\%, and 100.32\%) improvements in accuracy, precision, recall, and F1-score when using AI-Catcher in the DEV10 setup, respectively.

\begin{table}[h!]
\caption{Results of AI-generated text detection methods - DEV1.}
\label{dev1res}
\centering
\begin{tabular}{|l|c|c|c|c|}
\hline
\multicolumn{1}{|c|}{\textbf{Method}} & \textbf{Accuracy} & \textbf{Precision} & \textbf{Recall} & \textbf{F1-score} \\ \hline
LogReg                                & 0.785             & 0.769231           & 0.76087         & 0.765027          \\ \hline
RandF                                 & 0.795             & 0.765625           & 0.798913        & 0.781915          \\ \hline
SVM                                   & 0.6675            & 0.652695           & 0.592391        & 0.621083          \\ \hline
NB                                    & 0.4625            & 0.423645           & 0.467391        & 0.444444          \\ \hline
CNN                                   & 0.8525            & 0.822727           & 0.900498        & 0.859857          \\ \hline
MLP                                   & 0.6825            & 0.662281           & 0.751244        & 0.703963          \\ \hline
MLP2                                  & 0.66              & 0.627451           & 0.79602         & 0.701754          \\ \hline
MLP2-TP                               & 0.4975            & 0                  & 0               & 0                 \\ \hline
MLP5-TP                               & 0.5025            & 0.5025             & \textbf{1}               & 0.668885          \\ \hline

charCNN    &          0.792	& 0.907& 	0.871	& \textbf{0.888}         \\ \hline

XLNet & 0.825 &	0.7435	& 0.9951 &	0.8512  \\ \hline

ZeroGPT                               &      0.55             &               0.86207    &    0.124379             &     0.2174              \\ \hline	
GPTZero                               &       0.7975            &            \textbf{0.96154}        &    0.621891             &      0.755288             \\ \hline
Writer                                &       0.4925            &            0.428572        &   0.02986       &  0.055814       \\ \hline
AI-Catcher                            & \textbf{0.875}             & 0.864734           & 0.890547        & 0.8775             \\ \hline
\end{tabular}
\end{table}	

\begin{table}[h!]
\caption{Results of AI-generated text detection methods - DEV5 and DEV10.}
\label{dev510res}
\centering
\hspace*{-2cm}
\begin{tabular}{|l|cccc|cccc|}
\hline
\multicolumn{1}{|c|}{\multirow{2}{*}{\textbf{Method}}} & \multicolumn{4}{c|}{\textbf{DEV5}}                                                                                                          & \multicolumn{4}{c|}{\textbf{DEV10}}                                                                                                         \\ \cline{2-9} 
\multicolumn{1}{|c|}{}                                 & \multicolumn{1}{c|}{\textbf{Accuracy}} & \multicolumn{1}{c|}{\textbf{Precision}} & \multicolumn{1}{c|}{\textbf{Recall}} & \textbf{F1-score} & \multicolumn{1}{c|}{\textbf{Accuracy}} & \multicolumn{1}{c|}{\textbf{Precision}} & \multicolumn{1}{c|}{\textbf{Recall}} & \textbf{F1-score} \\ \hline
LogReg                                                 & \multicolumn{1}{c|}{0.8695}            & \multicolumn{1}{c|}{0.890827}           & \multicolumn{1}{c|}{0.843}           & 0.866254          & \multicolumn{1}{c|}{0.869}             & \multicolumn{1}{c|}{0.893301}           & \multicolumn{1}{c|}{0.839}           & 0.8653            \\ \hline
RandF                                                  & \multicolumn{1}{c|}{0.8775}            & \multicolumn{1}{c|}{0.881172}           & \multicolumn{1}{c|}{0.874}           & 0.877571          & \multicolumn{1}{c|}{0.876}             & \multicolumn{1}{c|}{0.875681}           & \multicolumn{1}{c|}{0.877}           & 0.87634           \\ \hline
SVM                                                    & \multicolumn{1}{c|}{0.8385}            & \multicolumn{1}{c|}{0.923756}           & \multicolumn{1}{c|}{0.739}           & 0.821113          & \multicolumn{1}{c|}{0.848}             & \multicolumn{1}{c|}{0.926471}           & \multicolumn{1}{c|}{0.756}           & 0.832599          \\ \hline
NB                                                     & \multicolumn{1}{c|}{0.6835}            & \multicolumn{1}{c|}{0.689623}           & \multicolumn{1}{c|}{0.666}           & 0.677606          & \multicolumn{1}{c|}{0.7055}            & \multicolumn{1}{c|}{0.711686}           & \multicolumn{1}{c|}{0.696}           & 0.703755          \\ \hline
CNN                                                    & \multicolumn{1}{c|}{0.9655}            & \multicolumn{1}{c|}{0.968775}           & \multicolumn{1}{c|}{0.961}           & {0.96479}           & \multicolumn{1}{c|}{0.978}             & \multicolumn{1}{c|}{0.974929}           & \multicolumn{1}{c|}{\textbf{0.983}}           & 0.978799          \\ \hline
MLP                                                    & \multicolumn{1}{c|}{0.6915}            & \multicolumn{1}{c|}{0.692309}           & \multicolumn{1}{c|}{0.708}           & 0.694887          & \multicolumn{1}{c|}{0.8085}            & \multicolumn{1}{c|}{0.79952}            & \multicolumn{1}{c|}{0.834}           & 0.81529           \\ \hline
MLP2                                                   & \multicolumn{1}{c|}{0.7075}            & \multicolumn{1}{c|}{0.715926}           & \multicolumn{1}{c|}{0.707}           & 0.705195          & \multicolumn{1}{c|}{0.8175}            & \multicolumn{1}{c|}{0.813293}           & \multicolumn{1}{c|}{0.837}           & 0.82315           \\ \hline
MLP2-TP                                                & \multicolumn{1}{c|}{0.697}             & \multicolumn{1}{c|}{0.717254}           & \multicolumn{1}{c|}{0.676}           & 0.680144          & \multicolumn{1}{c|}{0.5}               & \multicolumn{1}{c|}{0.15}               & \multicolumn{1}{c|}{0.3}             & 0.2               \\ \hline
MLP5-TP                                                & \multicolumn{1}{c|}{0.5}               & \multicolumn{1}{c|}{0.4}                & \multicolumn{1}{c|}{0.8}             & 0.533333          & \multicolumn{1}{c|}{0.5}               & \multicolumn{1}{c|}{0.15}               & \multicolumn{1}{c|}{0.3}             & 0.2               \\ \hline
AI-Catcher                                             & \multicolumn{1}{c|}{\textbf{0.968}}             & \multicolumn{1}{c|}{\textbf{0.965514}}           & \multicolumn{1}{c|}{\textbf{0.972}}           & \textbf{0.968646}         & \multicolumn{1}{c|}{\textbf{0.981}}            & \multicolumn{1}{c|}{\textbf{0.979674}}           & \multicolumn{1}{c|}{0.982}           & \textbf{0.980758}          \\ \hline
\end{tabular}
\end{table}

Table \ref{gptC} and Table \ref{humC} illustrate the precision, recall, and F1-score results obtained by AI-Catcher for the "ChatGPT" and "Human" classes, respectively. The results demonstrated the capability of AI-Catcher to detect Human-written and ChatGPT-generated text accurately. 

The second experiment focuses on comparing AI-Catcher with commercial detection tools. The results of the commercial detection tools are obtained by manual testing on the same 20\% testing dataset as utilized for AI-Catcher in the DEV1 setup. Table \ref{dev1res} illustrates the results of AI-Catcher and the commercial detection tools. On average, AI-Catcher improves the accuracy, precision, recall, and F1-score by  48.83\%, 30.68\%, 1180.54\%, and 597.34\%, respectively. 
	
\begin{table}[h!]
\caption{Results obtained by AI-Catcher for the ChatGPT class.}
\label{gptC}
\centering
\hspace*{-2cm}
\begin{tabular}{|c|c|c|c|c|c|c|}
\hline
\textbf{\begin{tabular}[c]{@{}c@{}}Experimental\\ Setup\end{tabular}} & \textbf{\begin{tabular}[c]{@{}c@{}}\#ChatGPT in\\ Testing Data\end{tabular}} & \textbf{\begin{tabular}[c]{@{}c@{}}\#Correct \\ Prediction\end{tabular}} & \textbf{\# Predictions} & \textbf{Precision} & \textbf{Recall} & \textbf{F1-score} \\ \hline
DEV1                                                                  & 201                                                                          & 179                                                                      & 207                     & 0.86474            & 0.89055         & 0.87746           \\ \hline
DEV5                                                                  & 1000                                                                         & 972                                                                      & 1008                    & 0.96429            & 0.972           & 0.96813           \\ \hline
DEV10                                                                 & 1000                                                                         & 982                                                                      & 1002                    & 0.98004            & 0.982           & 0.98102           \\ \hline
\end{tabular}
\end{table}

\begin{table}[h!]
\caption{Results obtained by AI-Catcher for the Human class.}
\label{humC}
\centering
\hspace*{-2cm}
\begin{tabular}{|c|c|c|c|c|c|c|}
\hline
\textbf{\begin{tabular}[c]{@{}c@{}}Experimental\\ Setup\end{tabular}} & \textbf{\begin{tabular}[c]{@{}c@{}}\#Human in\\ Testing Data\end{tabular}} & \textbf{\begin{tabular}[c]{@{}c@{}}\#Correct \\ Prediction\end{tabular}} & \textbf{\# Predictions} & \textbf{Precision} & \textbf{Recall} & \textbf{F1-score} \\ \hline
DEV1                                                                  & 199                                                                        & 171                                                                      & 193                     & 0.88601            & 0.8593          & 0.87245           \\ \hline
DEV5                                                                  & 1000                                                                       & 964                                                                      & 992                     & 0.97178            & 0.964           & 0.96788           \\ \hline
DEV10                                                                 & 1000                                                                       & 980                                                                      & 998                     & 0.98197            & 0.98            & 0.98099           \\ \hline
\end{tabular}
\end{table}

\subsection{Discussion and Implications}
\label{dis}

LLMs raise several ethical concerns for the scientific community. The emergence of AI-driven text generation tools transforms plagiarism into a more critical issue. The fabrication of research has severe consequences for the scientific community. AI-powered tools have the potential to generate scientific content that is indistinguishable from those written by a human. The core goal is to develop a novel AI-generated scientific text detection method, AI-Catcher.  AI-Catcher is a shared learning model trained on two types of features: (1) linguistic and statistical features; and (2) textual content. The classification model of AI-Catcher uses hidden patterns derived from MLP and CNN. 

The average accuracy results achieved by the proposed method is 94.14\%. On average, AI-Catcher improved accuracy by 38.8\% compared to alternative methods. The average precision results of AI-Catcher are 93.64\%  and 94.66\% for the "ChatGPT" and "Human" classes, respectively. The average precision improvements achieved by the proposed method is 59.7\%. Significant improvement in recall (i.e., 324.2\%) is demonstrated when using AI-Catcher. AI-Catcher achieved average recall results of 94.82\% and 93.45\% for the "ChatGPT" and "Human" classes, respectively. The F1-score results of the AI-Catcher for the "ChatGPT" and "Human" classes (94.23\% and 94.05\%) indicate the capability of the proposed method to capture the hidden patterns representing each class accurately. On average, AI-Catcher improved the F1-score by 190.08\% compared to alternative methods.

The implications of developing an AI-generated scientific textual content detection method are significant. Firstly, our work maintains the integrity and trustworthiness of academic and scientific literature. AI-Catcher assists in identifying and flagging AI-generated text that may be inaccurate, misleading, or fraudulent. Hence,
\begin{itemize}
\item Mitigate the risks associated with plagiarism and fraudulent academic publications.
\item Promote transparency and accountability in scientific research.  
\end{itemize}
 
Additionally, AI-Catcher has potential implications for education and academic publishers. In education, AI-Catcher could help professors and students avoid plagiarism and identify AI-generated content they may encounter in their research. Therefore, the adoption of our method could improve the quality of student work and promote academic integrity. For publishers, our approach could help editors identify AI-generated content submitted for publication to ensure that the published content is accurate, reliable, and original.

\section{Conclusion and Future Works}
\label{con}

In this research, we formulated the challenge of detecting ChatGPT-generated scientific text as a binary classification problem. AI-Catcher integrates two deep learning models, MLP and CNN, to learn high-level representation features for the human/ChatGPT classification task. The MLP extracts the feature representations of linguistic and statistical features. The CNN extracts high-level representations of the sequential patterns from the textual content. AI-Catcher uses a concatenation layer to fuse hidden patterns derived from MLP and CNN. 

AI-Catcher required collecting a new dataset (i.e., AIGTxt) to develop the classification model. AIGTxt is a new ChatGPT-generated scientific text dataset designed to enhance AI-generated text detection tools. The collected data involves three classes: Human-written, ChatGPT-generated, and Mixed text. AIGTxt contains 1000 records per class, collected from published academic articles across ten domains.

We conducted several experiments to evaluate the performance of the proposed model in terms of accuracy, precision, recall, and score. The comparative results showed that the AI-Catcher outperformed alternative methods. To the best of our knowledge, AI-Catcher is the first model that can distinguish between human-written and ChatGPT-generated scientific text. In addition, the AIGTxt dataset is the first public dataset that includes ChatGPT-generated scientific texts. In the future, the AI-Catcher can handle a multi-class problem by detecting the mixed text class. 

\section*{Funding}
Not applicable

\section*{Conflicts of interest}
No author associated with this paper has disclosed any potential or pertinent conflicts that may be perceived to have impending conflict with this work.

\section*{Ethics approval}
Not applicable

\section*{Consent to participate}
Not applicable

\section*{Consent for publication}
Not applicable

\section*{Availability of data and material}
The dataset used to build and evaluate the model is one of the contributions. The data will be available on a GitHub repository after the research is accepted for publication. 

\section*{Code availability}
The code will not be available to the public. 

\section*{Authors' contributions}
Bushra Alhijawi wrote the main manuscript text, prepared figures, developed the idea \& solution, and supervised the research group. Rawan Jarar implemented and tested the solution. Aseel AbuAlRub and Arwa Bader collected the dataset. All authors reviewed the manuscript text.

%========================================

% \bibliographystyle{apa}
\bibliography{ref}

\end{document}